%% file: shape_ml_2026_lnncs_main.tex
\documentclass[runningheads]{llncs}

\usepackage[T1]{fontenc}
\usepackage{graphicx}
\usepackage{multirow}

\begin{document}

\title{Mind the Gap: Mesh-Guided Repair of Broken Vessels}

% \titlerunning{Mesh-Guided Vessel Repair}

\author{
Gniewosz Drwiega\inst{1}\thanks{Corresponding author}\orcidID{0000-0002-1968-2238} \and
Wojciech Szymanski\inst{1,2}\orcidID{0009-0001-2043-9104} \and
Marek Wodzinski\inst{1,2}\orcidID{0000-0002-8076-6246}
}

\authorrunning{G. Drwiega et al.}

\institute{
Sano -- Centre for Computational Personalised Medicine International Research Foundation, Krakow, Poland\\
\and
AGH University of Krakow, Krakow, Poland
\email{\{g.drwiega,w.szymanski,m.wodzinski\}@sanoscience.org}
}

\maketitle

\begin{abstract}
Vessel segmentation is commonly optimized as voxel-wise classification, but small local errors can strongly disrupt vascular connectivity while having little effect on overlap scores. This is particularly problematic for downstream analyzes that rely on centerlines, branches, connected components, or graph structure. We propose a mesh-guided post-processing framework for repairing broken vessel segmentations produced by nnU-Net. For each predicted binary mask, a deformable template mesh is fitted to the mask surface in physical space and used as a case-specific geometric scaffold. The fitted mesh is not voxelized as the final segmentation; instead, it guides conservative reconnection of disconnected components by proposing or validating thin bridge candidates under foreground-growth constraints. We evaluated this approach in three vascular anatomies using AortaSeg24 and SEGA for the aorta, TopCoW for the Circle of Willis, and PARSE for the pulmonary arteries. Performance is measured using Dice, connected-component Dice (ccDice), and the Betti-0 number. Across these datasets, repair substantially improved connectivity while preserving overlap: Dice remained nearly unchanged, whereas ccDice increased from 0.596 to 0.992 for aorta, from 0.722 to 0.835 for TopCoW, and from 0.028 to 0.862 for PARSE. The FOMAML meta-initialization further accelerated the fitting per-case, supporting practical mesh-based repair of the vascular topology. These results suggest that explicit mesh representations can provide a useful geometric prior for correcting topological failures in otherwise accurate voxel segmentations.  

\keywords{Vessel segmentation \and Mesh deformation \and Topology \and Tubular Shapes}
\end{abstract}

\input{shape_ml_2026_introduction}
\input{shape_ml_2026_method}

\input{shape_ml_2026_results}
\input{shape_ml_2026_discussion_conclusion}

\begin{credits}
\subsubsection{\ackname}
This work was supported by the National Science Centre, Poland, under Grant ``MultiGeoMed'' No. 2024/55/D/ST6/02081. We gratefully acknowledge the Polish high-performance computing infrastructure PLGrid, HPC Center: ACK Cyfronet AGH, for providing computational resources and support within computational grant No. PLG/2025/018770.

\subsubsection{\discintname}
The authors have no competing interests to declare that are relevant to the content of this article.
\end{credits}

\bibliographystyle{splncs04}
\bibliography{shape_ml_2026_references}

\end{document}

%% file: shape_ml_2026_introduction.tex
\section{Introduction}

Medical image segmentation is often formulated as dense voxel classification \cite{ronneberger2015unet,isensee2021nnunet}, yet many target anatomies are more naturally described as geometric objects embedded in physical space \cite{heimann2009statisticalshape,bohlender2021shapeconstraint}. Organ boundaries, vascular trees, and anatomical loops are not only sets of foreground voxels. They have surfaces, centerlines, branches, connected components, and topological relationships that determine whether a segmentation is useful for measurement, planning, or modeling \cite{lesage2009vesselreview,antiga2008imagebased}. This perspective is especially relevant for vascular segmentation, where anatomical shape, branching structure, and connectivity are central to clinical and computational usability. Explicit shape representations, such as surfaces, meshes, or statistical shape models, can complement image-based learning by making anatomical geometry directly accessible to the model and to downstream analysis pipelines \cite{wickramasinghe2020voxel2mesh,kong2021meshdeformnet,bohlender2021shapeconstraint}. Large-scale resources such as MedShapeNet further facilitate the development and evaluation of methods based on explicit 3D anatomical shape representations \cite{li2023medshapenet}.

\begin{figure}
\centering
\includegraphics[width=\textwidth]{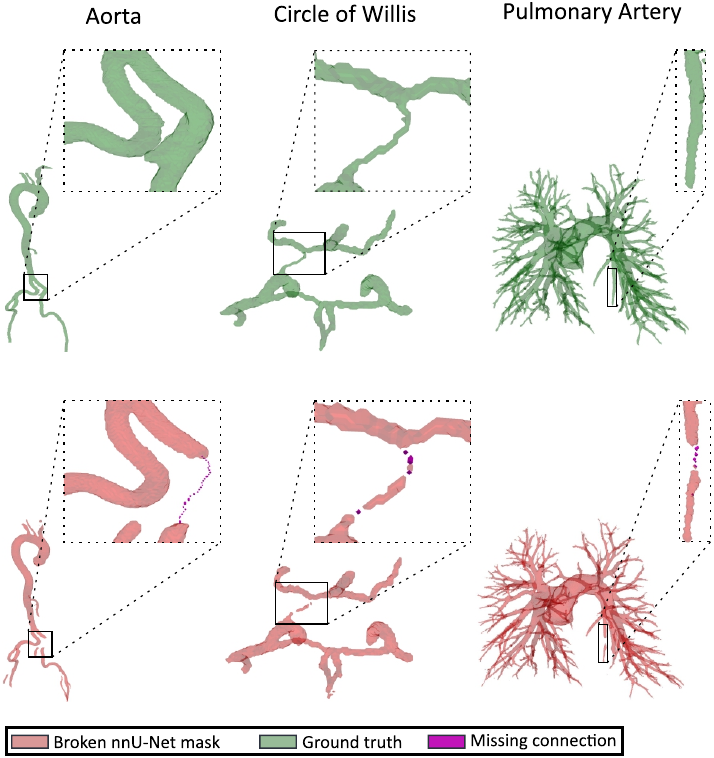}
\caption{Examples of broken-vessel failure modes across vascular anatomies. Top row: reference vessel structures. Bottom row: broken nnU-Net predictions, with missing local connections highlighted. Such small local errors can fragment otherwise accurate masks.}
\label{fig:broken_vessel_problem}
\end{figure}

The segmentation of vessels exposes this gap particularly clearly. Large vessels and vascular networks are thin, elongated, and strongly anisotropic. Their clinically meaningful properties depend on continuity and branching as much as on local overlap \cite{lesage2009vesselreview,shit2021cldice,decroocq2025benchmarking}. A small missing bridge can split a vessel into multiple components while changing only a tiny fraction of the volume (see Fig.~\ref{fig:broken_vessel_problem}). These errors are easy to underweight when optimizing or reporting voxel-wise overlap, but they matter for centerline extraction, branch labeling, diameter measurement, computational flow analysis, and any application that interprets the vascular structure as a connected object or graph \cite{pepe2020aorticdissection}. This has motivated shape-aware approaches that incorporate tubular geometry directly into feature extraction for coronary and cerebral vessel analysis \cite{wang2025tubular}. Recent vascular benchmarks such as AortaSeg24, SEGA, TopCoW, and PARSE further emphasize that vessel segmentation is not only a regional labeling task, but also a geometric and topological reconstruction problem \cite{imran2025aortaseg24,yang2025topcow,luo2023parse}.

Convolutional segmentation networks have made this problem much more tractable. U-Net established 
the encoder-decoder paradigm for biomedical segmentation, and nnU-Net demonstrated how robust configuration choices can turn this paradigm into a strong general-purpose baseline across many medical tasks \cite{ronneberger2015unet,isensee2021nnunet}. However, these methods still produce masks on a voxel grid and are commonly trained with losses that primarily reward local agreement with the reference annotation. For tubular structures, high Dice scores can therefore coexist with broken vessels, missing thin branches, isolated false-positive islands, or incorrect loops. Such errors may have limited impact on volumetric overlap while substantially changing the connectivity and topology of the predicted anatomy. This discrepancy motivates evaluation criteria and learning objectives that explicitly account for structure, rather than relying only on voxel-wise agreement.

The community has responded with topology-aware losses, skeleton-based objectives, and topology-specific evaluation measures. Persistent-homology losses penalize Betti-number errors \cite{hu2019topology}, while clDice and centerline-Cross Entropy encourage tubular continuity using skeleton or centerline information \cite{shit2021cldice,acebes2024clce}. On the evaluation side, ccDice extends the Dice principle from voxel overlap to spatially corresponding connected components, making it directly relevant for fragmented vessel predictions \cite{rouge2025ccdice}. Recent benchmarking further shows that overlap scores alone do not capture centerline continuity, component structure, and topological correctness \cite{decroocq2025benchmarking}. These works provide important training and evaluation signals, but they do not directly address post-processing of an already available voxel mask, where residual connectivity errors should be corrected while preserving the original prediction as much as possible.

Mesh representations provide a natural mechanism for such a correction. A mesh is a compact and explicit geometric scaffold: its vertices, edges, and faces encode adjacency, surface smoothness, and allowable paths in continuous physical space \cite{botsch2010polygonmesh,wickramasinghe2020voxel2mesh,kong2021meshdeformnet}. Prior work has used learned mesh deformation to reconstruct anatomical surfaces directly from volumetric images, including Voxel2Mesh and MeshDeformNet-style architectures \cite{wickramasinghe2020voxel2mesh,kong2021meshdeformnet}. Related work has also fitted parametric vessel models directly to segmentations, linking voxel, centerline-radius, and mesh representations through differentiable voxelization \cite{dima2026parametric}. Together, these methods illustrate why meshes are attractive for medical shape analysis: they can reduce staircase artifacts, preserve explicit connectivity, and expose geometry in a form that is easier to regularize than an unconstrained voxel map. However, most mesh-based segmentation methods use the mesh as the final anatomical representation. In vascular segmentation, extensive branching, variable caliber, fine-scale geometry, and substantial inter-patient variability make direct mesh-based replacement challenging. Mesh-deformation models must generalize across both global anatomical variation and small peripheral branches. If the fitted mesh is voxelized as a direct mask replacement, it may overfill the vessel volume or introduce anatomically implausible shortcuts.

Our contributions are fourfold. First, we propose a case-wise deformable-mesh fitting pipeline that transforms an existing nnU-Net vessel mask \cite{ronneberger2015unet,isensee2021nnunet} into an explicit, case-specific geometric scaffold. Second, we use the fitted mesh to guide conservative path-based reconnection of disconnected components rather than replacing the mask through full mesh voxelization, thereby limiting unnecessary foreground growth. Third, we introduce FOMAML meta-initialization to accelerate per-case mesh fitting by reducing the required number of optimization steps. Finally, we evaluate the proposed repair framework on four vascular datasets, AortaSeg24, SEGA, PARSE, and TopCoW, using overlap, component-level, and topological metrics that capture the geometric failure modes motivating this work.

%% file: shape_ml_2026_method.tex
\section{Method}

\subsection{Pipeline Overview}

\begin{figure}[!tbp]
\centering
\includegraphics[width=\textwidth]{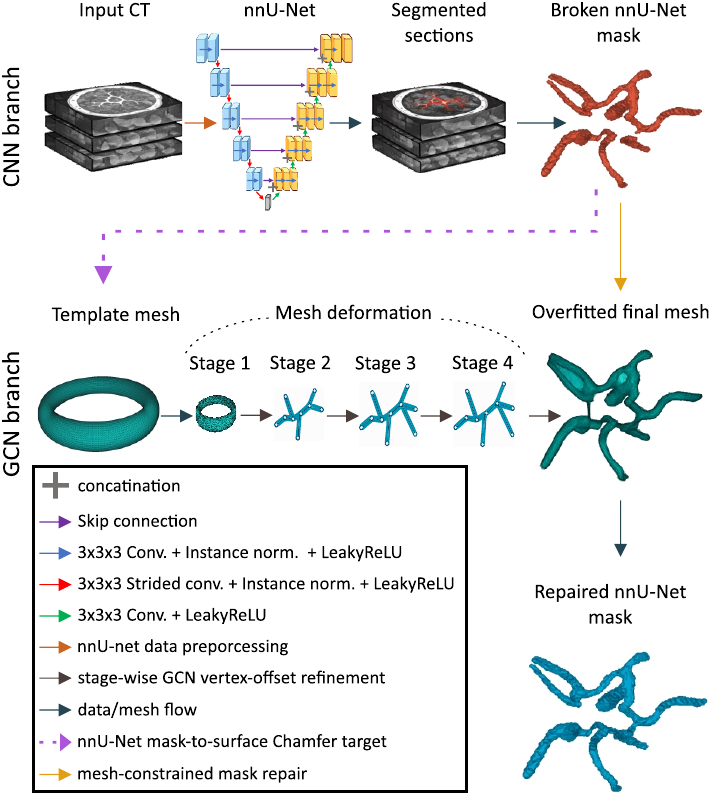}
\caption{Overview of the proposed mesh-guided nnU-Net repair pipeline.
A CT volume is first processed by a CNN segmentation branch based on nnU-Net, producing segmented vessel sections and an initial nnU-Net mask. When the predicted mask is fragmented or topologically inconsistent, its surface is extracted and used as the target for mesh fitting. Then, a fixed template mesh is passed through a GCN deformation branch. The mesh is progressively refined over multiple deformation stages, where each stage predicts bounded vertex offsets and forwards the updated mesh to the next refinement stage. The final overfitted mesh approximates the surface geometry of the nnU-Net prediction while preserving the template mesh connectivity. This fitted mesh is then used as a geometric prior for topology-aware mask repair, producing a repaired nnU-Net mask with improved connectivity.}
\label{fig:pipeline}
\end{figure}

The proposed method is a post-processing framework for repairing connectivity errors in vessel segmentations. Given a binary mask predicted by nnU-Net, we fit a deformable template mesh to the predicted vessel surface and use the fitted mesh as a case-specific geometric scaffold. The repaired output remains a voxel mask: the mesh is not used as the final segmentation, but only as a guide for targeted reconnection of disconnected components, as shown in Fig.~\ref{fig:pipeline}.

\subsection{Datasets and Input Masks}

We evaluate the proposed repair framework on vascular segmentation datasets covering different anatomical structures and topological regimes. For aortic anatomy, we use SEGA, which focuses on aortic vessel-tree segmentation from CT angiography with emphasis on robustness, visual quality, and downstream meshing, and AortaSeg24, which provides CTA volumes annotated for clinically relevant aortic branches and zones \cite{pepe2024sega,imran2025aortaseg24}. For cerebral vessels, we use TopCoW, a topology-aware Circle of Willis benchmark with voxel-level annotations of multiple CoW vessel components in CTA and MRA \cite{yang2025topcow}. For pulmonary vessels, we use PARSE, which targets multi-level pulmonary artery segmentation in CTPA scans \cite{luo2023parse}. Because the repair task concerns global vessel connectivity rather than multi-class anatomical labeling, all dataset-specific subregions or branch labels were merged into a single foreground vessel label before training and evaluation. 

\subsection{Mesh Fitting}

The mesh-fitting stage of the pipeline is illustrated in Fig.~\ref{fig:pipeline}. For each case, we fit a deformable template mesh to the surface of the predicted binary mask. Its associated NIfTI geometry, including voxel spacing, origin, and direction, is preserved so that all fitting operations are performed in physical coordinates. We extract the target surface with marching cubes and convert the resulting vertices from voxel-index coordinates to physical coordinates. To make optimization comparable across cases, the target surface is normalized by the center and scale of its physical bounding box.

The template mesh is deformed toward the normalized target surface using a graph-based mesh decoder. Starting from an anatomy-specific template, the decoder applies a sequence of deformation stages, where each stage predicts bounded vertex offsets conditioned on the current vertex positions and trainable per-vertex features \cite{wickramasinghe2020voxel2mesh,kong2021meshdeformnet}. In our experiments, we use a four-stage decoder and optimize its parameters separately for each case. We use a cylindrical template for aortic structures, a spherical template for PARSE, and a torus-like template for the Circle of Willis in TopCoW. The exact template resolution and optimization parameters are reported in the experimental protocol. 

The fitting loss combines Chamfer distance between sampled points on the deformed mesh and the target mask surface with geometric regularization terms, including edge-length, Laplacian, normal-consistency, and face-area regularization. These terms encourage the mesh to follow the predicted vessel surface while avoiding irregular or degenerate deformations. The fitting is performed independently for each predicted mask, rather than training a single mesh model to generalize across patients. For each case, the mesh with the best Chamfer score during optimization is mapped back to the original physical image space and used only as a local geometric prior for the subsequent repair step.

Unless stated otherwise, all mesh-fitting experiments used CUDA execution, random target-point sampling, 8192 target points for both the coarse and detail stages, 6000 fitting iterations, a detail-stage learning rate of 0.006, a Chamfer-loss weight of 1.0, decoder-stage maximum offsets of 0.35, 0.20, 0.10, and 0.05, and a decoder-stage auxiliary-loss weight of 0.1. For each dataset, per-case fitting was initialized from a mesh-model checkpoint obtained by overfitting one representative case. Dataset-specific template choices and regularization weights are reported in Table~\ref{tab:mesh_fit_hyperparameters}.

\input{Tables/training_hyperparameters_training_tables}

\subsection{Mesh-Guided Connectivity Repair}

After fitting, connectivity repair is performed conservatively. Rather than voxelizing the entire fitted mesh, the method identifies disconnected mask components and adds only thin bridge candidates that satisfy geometric validation criteria. To generate these candidate bridges, we consider two complementary strategies: one that follows the connectivity of the fitted mesh graph, and one that proposes local endpoint-to-endpoint bridges validated by the fitted mesh. In the mesh-graph variant, disconnected mask components are anchored to nearby vertices of the fitted mesh, and candidate bridges are obtained as shortest paths on the mesh graph. In the endpoint variant, the method instead proposes short local bridges between nearby component boundary points and uses the fitted mesh only for validation, requiring the bridge to be supported by nearby mesh vertices or face centroids. In both variants, a candidate bridge is rasterized as a thin tube and accepted only if it merges the target components while keeping the number of newly added voxels below a prescribed per-bridge foreground-growth limit.

Optional component filtering can remove distant false-positive mask components before mesh fitting and repair. After repair, a mesh-supported cleanup step can remove small components that remain poorly supported by the fitted mesh.

In our experiments, component-distance artifact filtering and mesh-supported cleanup were enabled for all datasets. Unless stated otherwise, repair used 26-connectivity, adaptive local bridge radii with a 6.0 mm local radius window, minimum component size of one voxel, post-repair mesh distance threshold of 2.0 mm, and minimum mesh-close fraction of 0.01. Dataset-specific repair variants and thresholds are reported in Table~\ref{tab:repair_hyperparameters}.

\input{Tables/repair_hyperparameters_table}

\subsection{FOMAML Meta-Initialization}

Meta-learned initialization has previously been used to accelerate case-specific medical shape reconstruction with implicit neural representations \cite{depaolis2025fast}. Motivated by this approach, we evaluate a first-order model-agnostic meta-learning (FOMAML) initialization for accelerating per-case deformable-mesh fitting \cite{finn2017maml,nichol2018firstorder}. Each training case is treated as a separate fitting task. For a given case, surface points sampled from the predicted mask are split into support and query sets. The mesh model is first adapted to the support points using several inner optimization steps, and the shared initialization is then updated using the query loss after adaptation. Following the first-order approximation, the meta-update does not backpropagate through the full inner optimization trajectory. At inference time, the learned initialization is used as a warm start for case-specific mesh fitting, with the goal of reducing optimization time and improving fitting stability.

\subsection{Evaluation Metrics}

We evaluate the repaired masks using both voxel-wise segmentation accuracy and topology-oriented measures. Dice similarity coefficient is used to quantify foreground overlap with the reference segmentation. To assess connected-component consistency, we report ccDice, a topology-aware Dice score that extends the Dice principle from individual voxels to connected components and accounts for their spatial correspondence \cite{rouge2025ccdice}. We also report the Betti-0 number, corresponding to the number of connected components in the binary vessel mask, as a direct measure of fragmentation.

Although additional overlap, skeleton, and component-level metrics were computed during analysis, we focus the main results on Dice, ccDice, and Betti-0 because they summarize the central trade-off addressed by the proposed method: preserving voxel-wise segmentation quality while improving vessel connectivity.

\subsection{Experimental Protocol}

The baseline for all experiments is the raw nnU-Net binary prediction before mesh-guided repair. nnU-Net models were trained using the 3D full-resolution configuration with the default \texttt{nnUNetTrainer}. Training was performed on the Helios PLGrid GPU cluster using one GPU per job, 16 CPU cores, 120 GB RAM, and a 48 h wall-time limit. As summarized in Fig.~\ref{fig:pipeline}, the input to the repair stage is a hard binary vessel mask predicted by the nnU-Net 3D full-resolution configuration. The raw nnU-Net prediction is used as the baseline, and the proposed method is applied as a post-processing step to the same mask. We use five-fold outer evaluation protocol, ensuring that each repaired case was segmented by a model that did not see it during training. The repair stage does not use image intensities or nnU-Net probability maps.

Mesh fitting and connectivity repair were then run as a post-processing step on the predicted binary masks. These experiments were performed on a local workstation with an Intel Core Ultra 9 275HX CPU, 32 GB RAM, and an NVIDIA GeForce RTX 5090 GPU with 24 GB memory. The mesh-fitting hyperparameters were selected by grid-search experiments on the Helios cluster and then fixed for the reported out-of-fold evaluation. All post-processing experiments were implemented in Python using PyTorch for mesh deformation, SimpleITK for NIfTI image I/O and geometry handling, scikit-image for surface extraction, SciPy for connected-component and nearest-neighbor operations, and nnU-Net v2 for baseline segmentation. The source code is available in the accompanying repository \cite{drwiega2026code}.

%% file: Tables/training_hyperparameters_training_tables.tex
\begin{table}
\caption{Dataset-specific mesh-fitting hyperparameters.}
\label{tab:mesh_fit_hyperparameters}
\centering
\small
\setlength{\tabcolsep}{4pt}
\begin{tabular}{@{}lllll@{}}
\hline
Dataset & Template & Edge & Lap. & Normal / Area \\
\hline
Aorta Unified & Cylinder, 10k & 0.03125 & 1.25 &
$2.5{\times}10^{-4}$ / $2.5{\times}10^{-5}$ \\
TopCoW & Torus-like, 10k & 0.000625 & 0.025 &
$2.5{\times}10^{-4}$ / $2.5{\times}10^{-5}$ \\
PARSE & Sphere, 20k & 0.1 & 5.0 &
$1.0{\times}10^{-3}$ / $5.0{\times}10^{-4}$ \\
\hline
\end{tabular}
\end{table}

%% file: Tables/repair_hyperparameters_table.tex
\begin{table}
\caption{Dataset-specific mask-repair hyperparameters.}
\label{tab:repair_hyperparameters}
\centering
\small
\setlength{\tabcolsep}{3pt}
\begin{tabular}{@{}p{0.25\textwidth}p{0.22\textwidth}p{0.22\textwidth}p{0.22\textwidth}@{}}
\hline
Parameter & Aorta Unified & TopCoW & PARSE \\
\hline
Repair variant & Mesh-graph path & Mesh-graph path & Endpoint + mesh validation \\
Artifact keep/remove (mm) & 57.0 / 57.5 & 57.0 / 57.5 & 20.0 / 25.0 \\
Bridge anchor/support & Anchor 3.0 mm & Anchor 3.0 mm & Gap 12.0 mm; support 3.0 mm; fraction 0.50 \\
Tube radius base/min/max (mm) & 1.0 / 2.0 / 5.0 & 0.25 / 0.35 / 1.0 & 0.4 / 0.7 / 1.8 \\
Adaptive radius percentile/scale & 80 / 1.0 & 80 / 0.7 & 70 / 0.8 \\
Max added foreground fraction & 0.03 & 0.15 & 0.06 \\
Cleanup max removed voxels & 1000 & 1000 & 200 \\
\hline
\end{tabular}
\end{table}

%% file: shape_ml_2026_results.tex
\section{Results}

\subsection{Visual Assessment of Vessel Reconnection}

\begin{figure}[!tbp]
\centering
\includegraphics[width=0.95\textwidth]{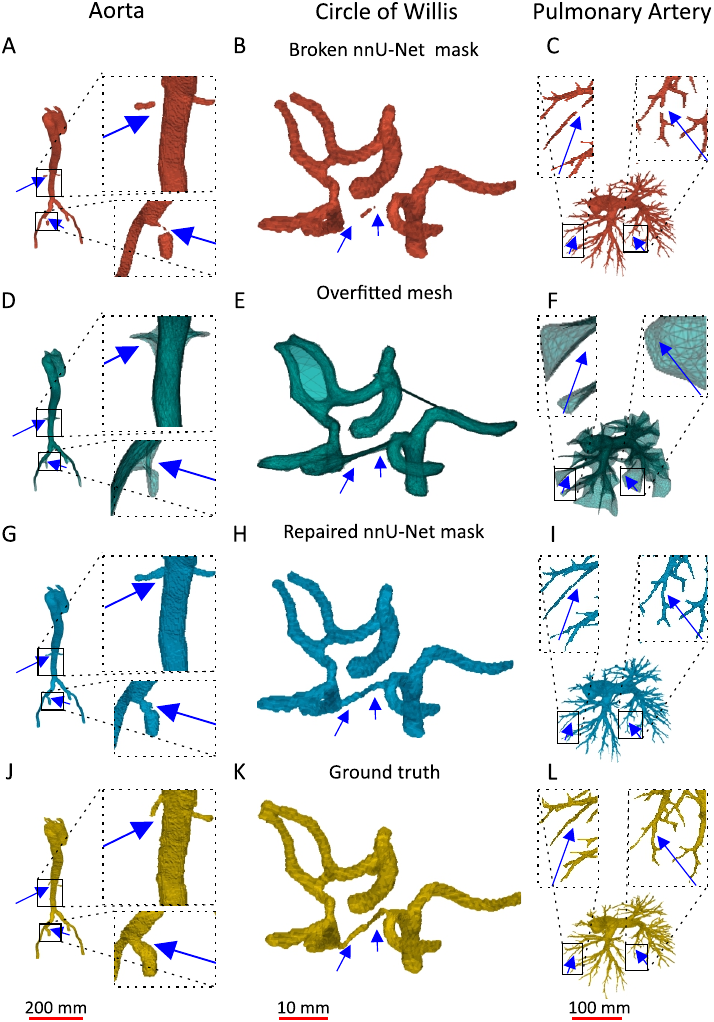}
\caption{Qualitative examples of mesh-guided repair across vascular anatomies. The method reconnects fragmented vessel components using local mesh-guided bridges while preserving the original mask away from the repaired regions. Blue arrows indicate broken vessel regions repaired by the proposed method.}
\label{fig:qualitative_results}
\end{figure}

Qualitative examples are shown in Fig.~\ref{fig:qualitative_results}. Broken-connectivity errors were observed across all evaluated anatomies, but with different visual patterns. In the aortic cases, errors typically appeared as small missing bridges along otherwise well-segmented large vessels or near branch regions (Fig.~\ref{fig:qualitative_results}A). The TopCoW cases showed the most frequent local discontinuities in the posterior communicating artery region of the Circle of Willis (Fig.~\ref{fig:qualitative_results}B). In PARSE, the pulmonary arterial tree exhibited severe fragmentation, especially in the thinner peripheral branches (Fig.~\ref{fig:qualitative_results}C).

Across these examples, mesh-guided repair added localized connector regions rather than replacing the full mask by a voxelized mesh. The repaired masks therefore preserved the original nnU-Net segmentation in most regions while reconnecting components that were geometrically supported by the fitted mesh (Fig.~\ref{fig:qualitative_results}D-L).

\subsection{Effect on Overlap and Connectivity}

The main quantitative results are summarized in Table~\ref{tab:main_results}. For the aortic cohort, including SEGA and AortaSeg cases, Dice remained essentially unchanged after repair, changing from $0.934 \pm 0.043$ for the raw mask to $0.934 \pm 0.043$ after repair. In contrast, ccDice increased from $0.596 \pm 0.275$ to $0.992 \pm 0.057$, and the Betti-0 number decreased from $3.35 \pm 2.61$ to $1.01 \pm 0.12$.

A similar pattern was observed for TopCoW. Dice changed only slightly, from $0.870 \pm 0.055$ to $0.867 \pm 0.054$, while ccDice improved from $0.722 \pm 0.247$ to $0.835 \pm 0.195$. The Betti-0 number decreased from $2.58 \pm 1.14$ to $1.02 \pm 0.13$, indicating that most repaired predictions became nearly single-component structures.

The strongest connectivity effect was observed for PARSE, where raw pulmonary-artery masks were highly fragmented. Dice remained stable, changing from $0.877 \pm 0.036$ to $0.876 \pm 0.036$, while ccDice increased from $0.028 \pm 0.012$ to $0.862 \pm 0.211$. The Betti-0 number decreased from $87.46 \pm 45.92$ to $1.56 \pm 1.06$. The false-branch fraction remained nearly unchanged for the aortic cohort ($0.056 \pm 0.009$ to $0.055 \pm 0.009$) and PARSE ($0.130 \pm 0.020$ to $0.132 \pm 0.020$), but increased for TopCoW from $0.040 \pm 0.006$ to $0.063 \pm 0.007$, indicating a dataset-specific trade-off between restoring connectivity and avoiding spurious branches. These results support the central motivation of the method: large connectivity improvements can be achieved with negligible change in voxel-wise overlap.

\input{Tables/results_main_table_ci_multirow}

\begin{figure}[!tbp]
\centering
\includegraphics[width=0.95\textwidth]{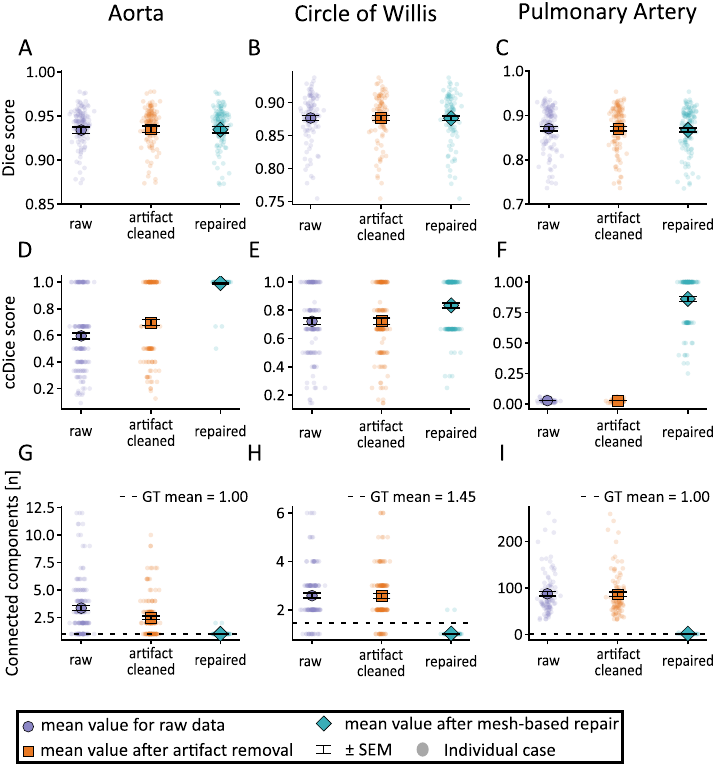}
\caption{Quantitative summary of repair effects and mesh-fitting convergence. Mesh-guided repair improves connectivity-oriented metrics while preserving Dice, and meta-initialization accelerates per-case mesh fitting.}
\label{fig:quantitative_results}
\end{figure}

\subsection{Effect of FOMAML Meta-Initialization}

\begin{figure}[!tbp]
\centering
\includegraphics[width=0.85\textwidth]{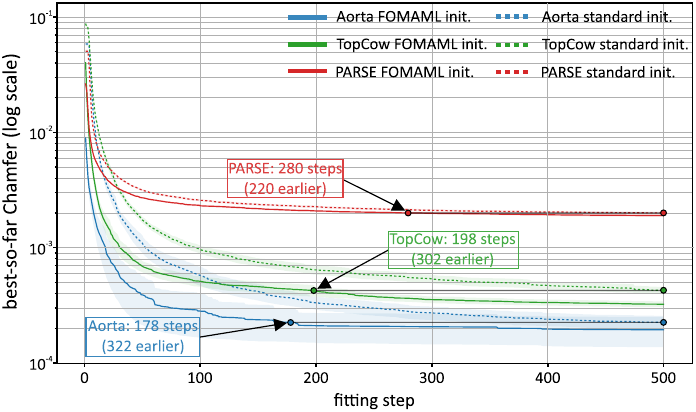}
\caption{Effect of FOMAML meta-initialization on per-case mesh-fitting convergence for 25 held-out TopCoW cases. Lines show mean Chamfer loss across cases, and shaded regions indicate $\pm$ SEM. The black horizontal line indicates a matched Chamfer-distance level, allowing comparison of the number of optimization steps required by FOMAML and standard initialization to reach the same reconstruction error.}
\label{fig:fomaml_convergence}
\end{figure}

One might expect fitting a mesh independently to each predicted mask to introduce substantial computational overhead during inference. To reduce this overhead, we use FOMAML to learn an initialization that can adapt to a new case in fewer optimization steps. Fig.~\ref{fig:fomaml_convergence} compares Chamfer-loss convergence between standard and FOMAML initialization on 25 held-out TopCoW cases. FOMAML achieved a lower Chamfer loss at every evaluated checkpoint and outperformed the standard initialization in all 25 cases, demonstrating faster and more consistent per-case mesh fitting.

The largest benefit was observed in the early optimization phase, indicating that FOMAML provides a better starting point for case-wise fitting. After 100 fitting iterations, mean Chamfer decreased from $7.79{\times}10^{-4}$ with standard initialization to $4.99{\times}10^{-4}$ with FOMAML. After 500 iterations, the corresponding values were $4.43{\times}10^{-4}$ and $3.39{\times}10^{-4}$. This suggests that FOMAML can reduce the number of iterations needed to reach a given fitting quality, providing a useful warm start for case-specific mesh optimization.

Overall, the results show that the proposed repair strategy mainly affects connectivity rather than regional overlap. This behavior is desirable for broken-vessel repair: the method improves component structure and connectedness while preserving the strong voxel-wise segmentation performance of nnU-Net.

%% file: Tables/results_main_table_ci_multirow.tex
\begin{table}
\caption{Segmentation and connectivity metrics before and after repair. Values are reported as mean $\pm$ 95\% confidence interval. Best values within each dataset are shown in bold. FB denotes false-branch fraction.}
\label{tab:main_results}
\centering
\scriptsize
\setlength{\tabcolsep}{3pt}
\begin{tabular}{@{}llcccc@{}}
\hline
Dataset & Mask & Dice $\uparrow$ & ccDice $\uparrow$ & $\beta_0$ $\downarrow$ & FB $\downarrow$ \\
\hline
\multirow{3}{*}{Aorta ($n=145$)}
& Raw      & $0.934 \pm 0.007$ & $0.596 \pm 0.045$ & $3.35 \pm 0.43$ & $0.056 \pm 0.009$ \\
& Filtered & $\mathbf{0.935 \pm 0.007}$ & $0.695 \pm 0.043$ & $2.47 \pm 0.30$ & $\mathbf{0.044 \pm 0.008}$ \\
& Repaired & $0.934 \pm 0.007$ & $\mathbf{0.992 \pm 0.009}$ & $\mathbf{1.01 \pm 0.02}$ & $0.055 \pm 0.009$ \\
\hline
\multirow{3}{*}{TopCoW ($n=125$)}
& Raw      & $\mathbf{0.870 \pm 0.010}$ & $0.722 \pm 0.043$ & $2.58 \pm 0.20$ & $\mathbf{0.040 \pm 0.006}$ \\
& Filtered & $\mathbf{0.870 \pm 0.010}$ & $0.723 \pm 0.043$ & $2.58 \pm 0.20$ & $\mathbf{0.040 \pm 0.006}$ \\
& Repaired & $0.867 \pm 0.010$ & $\mathbf{0.835 \pm 0.034}$ & $\mathbf{1.02 \pm 0.02}$ & $0.063 \pm 0.007$ \\
\hline
\multirow{3}{*}{PARSE ($n=100$)}
& Raw      & $\mathbf{0.877 \pm 0.007}$ & $0.028 \pm 0.002$ & $87.46 \pm 9.00$ & $0.130 \pm 0.020$ \\
& Filtered & $\mathbf{0.877 \pm 0.007}$ & $0.028 \pm 0.002$ & $86.56 \pm 8.97$ & $\mathbf{0.129 \pm 0.020}$ \\
& Repaired & $0.876 \pm 0.007$ & $\mathbf{0.862 \pm 0.041}$ & $\mathbf{1.56 \pm 0.21}$ & $0.132 \pm 0.020$ \\
\hline
\end{tabular}
\end{table}

%% file: shape_ml_2026_discussion_conclusion.tex
\section{Discussion}
\subsection{Mesh-Guided Topology Repair}
The results indicate that mesh representations can be useful as case-specific geometric priors for repairing vessel topology. Instead of replacing a voxel segmentation with a fully voxelized mesh, the proposed method uses the fitted mesh only to guide local reconnection. This distinction is important: the repaired output remains close to the original nnU-Net prediction, while the mesh provides additional geometric structure for identifying plausible paths between fragmented components. Because the mesh is fitted independently for each case, the method does not require a single mesh model to generalize directly across all patients and vascular anatomies. This per-case optimization is especially useful for vessels, where branching patterns, caliber, and topology vary substantially between cases. Across datasets, the main effect of the method was therefore observed in connectivity-sensitive metrics, whereas Dice changed only marginally.

The behavior was most consistent for aortic anatomy, where the target structure is comparatively large and the main connectivity errors are often localized. TopCoW and PARSE introduce more challenging topological settings. In the Circle of Willis, small vessels, variable anatomy, and clinically important communicating arteries mean that a local discontinuity can strongly affect graph connectivity. Moreover, some TopCoW ground-truth masks represent anatomically incomplete Circle of Willis variants; for example, absence of both posterior communicating arteries may leave the anterior and posterior circulations disconnected in the binary vascular mask \cite{yang2025topcow}. PARSE is also demanding because pulmonary arteries form a dense, highly branching tree with many peripheral vessels. In these cases, mesh fitting and bridge selection must handle substantially more complex geometry than in the aorta.

The FOMAML meta-initialization further improves the practicality of this per-case strategy. In the aortic setting, nnU-Net inference required 44.17 s per case, while FOMAML-initialized mesh fitting with approximately 300 optimization steps followed by post-repair required about 30 s per case. Thus, the additional geometric repair stage remains comparable to the original segmentation inference time while providing a targeted improvement in connectivity.

\subsection{Limitations}
Several limitations remain. First, the fitted mesh can support an incorrect bridge if the mask contains misleading false-positive components or if the mesh deformation follows an anatomically implausible shortcut. Second, although optional component filtering and mesh-supported cleanup reduce the influence of distant false positives, they do not fully solve cases where false positives are close to true vessels. Third, the current repair stage uses hard binary masks and does not exploit image intensities or nnU-Net probability maps, which could help distinguish plausible missing vessels from unsupported bridges. Finally, the method relies on dataset-specific templates and hyperparameters, which may need adjustment for new vascular territories.
\subsection{Future Work}
Future work will focus on making the repair step more adaptive. Learned bridge scoring could replace hand-designed acceptance rules, while anatomical constraints could reduce implausible shortcuts in highly variable structures such as the Circle of Willis and pulmonary vasculature. Incorporating probability maps or image-based evidence may also improve bridge validation, especially in ambiguous peripheral regions. We also plan to conduct computational hemodynamic experiments to quantify how small topological corrections affect simulated blood flow and other downstream functional measurements. More broadly, combining per-case mesh fitting with learned topology-aware repair may provide a stronger framework for correcting vascular segmentation without retraining the original segmentation model.

\section{Conclusion}

We presented a mesh-guided post-processing method for repairing broken vessel segmentations. By fitting a deformable mesh to each predicted mask and using it as a local geometric scaffold, the method improves vascular connectivity while preserving the original voxel-wise segmentation. With FOMAML initialization, the per-case repair remains practical in runtime while avoiding the need for a globally generalizing mesh prediction model. The results support the use of mesh representations as practical geometric priors for topology repair in vessel segmentation.